\documentclass[twoside,11pt]{article}

\usepackage[accepted]{melba}

\usepackage{mwe} 

\usepackage{amsmath,amsfonts}

\melbaheading{2022:014}{https://www.melba-journal.org/papers/2022:014.html}{2022}{1-27}{12/2021}{05/2022}{Valvano, Leo, Tsaftaris}{MICCAI 2021 Workshops Ominibus -- DART}{Ziyue Xu, Konstantinos Kamnitsas}

\ShortHeadings{Re-using Adversarial Mask Discriminators for Test-time Training}{Valvano, Leo and Tsaftaris}
\firstpageno{1}

\usepackage{amsmath}
\usepackage{xparse}
\usepackage{enumitem}
\usepackage{amssymb}
\usepackage{todonotes}
\usepackage{pifont}

\usepackage{url}
\usepackage{xurl}  
\usepackage{hyperref}

\makeatletter
\g@addto@macro{\UrlBreaks}{\UrlOrds}
\makeatother

\usepackage{color, colortbl}  
\usepackage{soul} 

\usepackage{booktabs}
\usepackage{multirow}  
\usepackage{rotating}  
\usepackage{placeins}  

\usepackage{algorithmic} 
\usepackage{amsmath}
\usepackage{dsfont}
\usepackage{upgreek}  
\usepackage{mathtools}

\ExplSyntaxOn
\NewDocumentCommand{\ucgreek}{m}
 {\str_case:nn { #1 } {
    {A}{\mathrm{A}} {B}{\mathrm{B}} {C}{\Sigma} {D}{\Delta} {E}{\mathrm{E}} 
    {F}{\Phi} {G}{\Gamma} {H}{\mathrm{H}} {I}{\mathrm{I}} {J}{\Theta} {K}{\mathrm{K}} 
    {L}{\Lambda} {M}{\mathrm{M}} {N}{\mathrm{N}} {O}{\mathrm{O}} {P}{\Pi}
    {Q}{\mathrm{X}} {R}{\mathrm{P}} {S}{\Sigma} {T}{\mathrm{T}} {U}{\Upsilon} 
    {W}{\Omega} {X}{\Xi} {Y}{\Psi} {Z}{\mathrm{Z}}
}}
\NewDocumentCommand{\lcgreek}{m}
 {\str_case:nn { #1 }
   {{a}{\alpha} {b}{\beta} {c}{\varsigma} {d}{\delta} 
    {e}{\varepsilon} {f}{\varphi} {g}{\gamma} {h}{\eta} {i}{\iota}
    {j}{\vartheta} {k}{\kappa} {l}{\lambda} {m}{\mu} {n}{\nu} {o}{o}
    {p}{\pi} {q}{\chi} {r}{\rho} {s}{\sigma} {t}{\tau} {u}{\upsilon} 
    {w}{\omega} {x}{\xi} {y}{\psi} {z}{\zeta}
}}
\ExplSyntaxOff

\newcommand{\mathimage}[1]{\mathbf{#1}}

\newcommand{\mathscal}[1]{\lowercase{\textit{#1}}}
\newcommand{\mathtensor}[1]{\mathrm{\uppercase{#1}}}
\newcommand{\mathdistrib}[1]{\mathcal{#1}}
\newcommand{\mathfunc}[1]{\uppercase{\ucgreek{#1}}} 
\newcommand{\norm}[1]{\left\lVert#1\right\rVert} 

\usepackage{pifont}

\newcommand{\xmark}{\ding{55}}

\makeatletter
\newcommand*{\resetnamehighlights}{\let\nhcbx@highlightlist\@empty}
\resetnamehighlights

\newcommand*{\highlightnames}{%
  \@ifstar{\resetnamehighlights\highlightnames@add}{\highlightnames@add}}

\newcommand*{\highlightnames@add@inner}[2]{%
  \listeadd{#1}{\the\numexpr#2\relax}}

\newcommand*{\highlightnames@add}{%
  \forcsvlist{\highlightnames@add@inner\nhcbx@highlightlist}}

\newcommand*{\mkbibhighlightnthname}[1]{%
  \xifinlist{\the\value{listcount}}{\nhcbx@highlightlist}
    {\mkbibbold{#1}}
    {#1}}
\makeatother

\usepackage[normalem]{ulem}

\definecolor{lightcyan}{HTML}{DAE8FC}
\definecolor{lightyellow}{RGB}{255,253,194}
\definecolor{lightpink}{RGB}{238,202,224}
\definecolor{lightgreen}{RGB}{175,217,141}
\newcommand{\hltext}[2][lightcyan]{{%
    \colorlet{foo}{#1}%
    \sethlcolor{foo}\hl{#2}}%
}

\definecolor{lightgray}{rgb}{0.92,0.92,0.92}
\newcolumntype{g}{>{\columncolor{lightgray}}c}
\newcolumntype{?}{!{\vrule width 1.5pt}}  
\newcommand{\hcell}[1]{\cellcolor[HTML]{DAE8FC}{#1}} 

\newcommand{\projectpageTestTimeTraining}{\url{https://vios-s.github.io/adversarial-test-time-training}} 

\newlist{todolist}{itemize}{2}
\setlist[todolist]{label=$\square$}

\title{Re-using Adversarial Mask Discriminators for Test-time Training under Distribution Shifts}
\author{\name Gabriele~Valvano \email gabriele.valvano@imtlucca.it \\
        \addr IMT School for Advanced Studies Lucca, Lucca LU, Italy \\
        \addr School of Engineering, University of Edinburgh, Edinburgh, UK
        \AND
        \name Andrea~Leo \email andrea.leo@unipi.it \\
        \addr 
        University of Pisa, Pisa, Italy
        \AND
        \name Sotirios~A.~Tsaftaris \email S.Tsaftaris@ed.ac.uk \\
        \addr School of Engineering, University of Edinburgh, Edinburgh, UK
}

\begin{document}

\graphicspath{{figures/}}

\maketitle

\begin{abstract}
Thanks to their ability to learn flexible data-driven losses, Generative Adversarial Networks (GANs) are an integral part of many semi- and weakly-supervised methods for medical image segmentation. GANs jointly optimise a generator and an adversarial discriminator on a set of training data. After training is complete, the discriminator is usually discarded, and only the generator is used for inference. But should we discard discriminators? In this work, we argue that training stable discriminators produces expressive loss functions that we can re-use at inference to detect and \textit{correct} segmentation mistakes. First, we identify key challenges and suggest possible solutions to make discriminators re-usable at inference. Then, we show that we can combine discriminators with image reconstruction costs (via decoders) to endow a causal perspective to test-time training and further improve the model. Our method is simple and improves the test-time performance of pre-trained GANs. Moreover, we show that it is compatible with standard post-processing techniques and it has the potential to be used for Online Continual Learning. With our work, we open new research avenues for re-using adversarial discriminators at inference. %
Our code is available at \projectpageTestTimeTraining.
\end{abstract}


\section{Introduction}

Generative Adversarial Networks (GANs, \cite{goodfellow2014generative}) have recently shown astonishing performance in semi-supervised and weakly-supervised segmentation learning \citep{cheplygina2019not,tajbakhsh2020embracing}, providing training signals when labels are sparse or missing.
GANs jointly optimise two networks in the adversarial setup: one to solve an image generation task (generator) and the other to distinguish between real images and the generated ones (discriminator or critic). 
In image segmentation, the generator is named segmentor and, conditioned on an input image, learns to predict correct segmentation masks. Meanwhile, the discriminator learns a data-driven shape prior and penalise the segmentor when it produces unrealistic predictions.
Once training is complete, the discriminator is discarded, and the segmentor is used for inference. 

Unfortunately, segmentors may decrease their test-time performance when the test data fall outside the training data distribution. Often, mistakes are easy to detect by the human eye as they appear as holes inside of the predicted masks or scattered false positives. Would a mask discriminator, trained to learn a shape prior, be able to spot such mistakes?

\begin{figure}[t]
    \centering
    \includegraphics[width=0.65\linewidth]{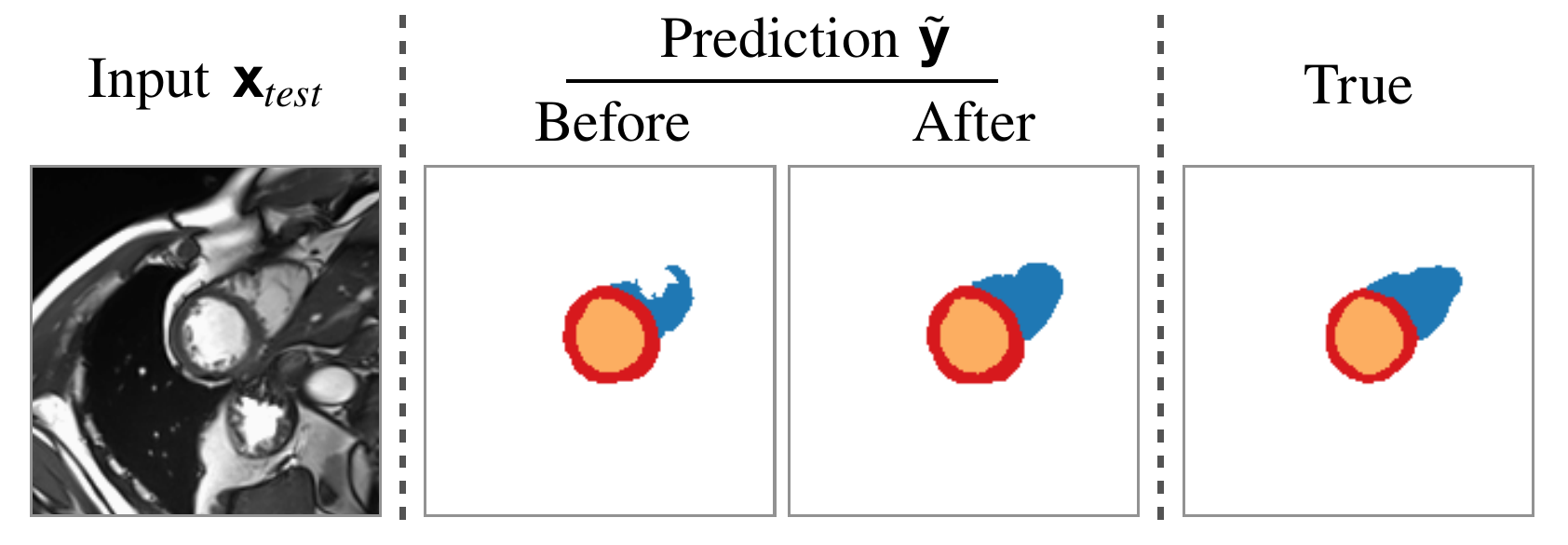}
    \caption{After training a GAN for semantic segmentation, whenever a test image falls outside the training data distribution, the segmentor may underperform and produce unrealistic predictions. Our work suggests re-using already optimised adversarial discriminators to tune the segmentor predictions on the individual test images until the predicted mask satisfies the learned shape prior.}
    \label{ch8:fig:abstract}
\end{figure} 
In this work, we introduce a simple mechanism to detect and correct such segmentation errors re-using components already developed during training. 
We adopt the emerging paradigm of Test-time Training: a term coined by \cite{sun2020test}. Test-time Training fine-tunes pre-trained networks on the individual test samples without any additional supervision, resulting in Variable Decision Boundary models \citep{sun2020test, wang2021tent, karani2021test, zhang2021memo, bartler2021mt3}. We present strategies to allow for the recycling of adversarial mask discriminators to make them useful at inference.  
Inspired by \cite{asano2019critical} showing that it is possible to train the early layers of a CNN with just a single image, we propose fine-tuning these layers on each test sample until the segmentor prediction satisfies the learned adversarial shape prior. 
We report an example of the benefits of our method when applied on initially erroneous predictions in Fig.~\ref{ch8:fig:abstract}, and summarise our \textbf{contributions} as follows:
\begin{itemize}
    \item To the best of our knowledge, our work is the first attempt to re-use adversarial mask discriminators at inference to \textit{detect} and \textit{correct} segmentation mistakes.
    \item We define specific assumptions that make discriminators still useful after training, and present possible solutions to satisfy them.
    \item We investigate the possibility to further improve test-time segmentation through causal learning, where we complement the discriminator with an image reconstructor.
    \item We explore various learning scenarios and report consistent performance increase on multiple medical datasets. 
\end{itemize}

This paper extends our previous publication \citep{valvano2021advttt} by: i) including a mechanism to set the number of iterations at test-time automatically; ii) evaluating our method on three additional datasets and in different training scenarios; iii) investigating if reconstruction losses can endow the model a causal perspective to improve model performance and inference speed; iv) analysing the method's compatibility with post-processing techniques; v) including experiments in continual learning settings; and vi) including additional ablations. 

\section{Related Work}

\subsection{Learning from Test Samples}
In our work, we use a discriminator to tune a segmentor on the individual \textit{test} images until it predicts realistic masks. The idea of unsupervised fine-tuning of a model on the test data has been recently introduced by \cite{sun2020test} and termed Test-time Training (TTT). 
To optimise a model on the training set, TTT suggests jointly minimising a supervised and an auxiliary self-supervised loss, such as predicting the rotation angle of a given image. 
After training, TTT uses the auxiliary task to fine-tune the model on the individual test samples and adapts to potential distribution shifts without the need for supervision.
Unfortunately, Sun et al. only test their method ``simulating" domain shifts through hand-crafted image corruptions, such as noise and blurring, and do not investigate if TTT can also improve semantic segmentation. Moreover, despite the success of TTT in image classification tasks, designing a well-suited auxiliary task is non-trivial \citep{sun2020test}. 
For instance, predicting rotation angles may be non-optimal in medical imaging, where images have different acquisition geometries. 

After this seminal work, \cite{wang2021tent} proposed tuning an adaptor network to minimise the prediction entropy on a test set. Unfortunately, this method needs access to the \textit{entire} test set for fine-tuning. Hence, \cite{zhang2021memo} proposed to focus on a single test point, minimising the marginal entropy of the model predictions under a set of data augmentations.
Unfortunately, neural networks are well known for making low-entropy overly-confident predictions \citep{guo2017calibration}, and minimising segmentation entropy could be sub-optimal. Moreover, the quality of the results also depends on making a good choice of augmentations \cite{zhang2021memo}. 

Recently, \cite{karani2021test} extended TTT to semantic segmentation by proposing Test-time Adaptable Neural Networks (TTANN). At first, TTANN learns a data-driven shape prior by pre-training a mask Denoising Autoencoder (DAE). At inference, the DAE auto-encodes the masks generated by a segmentor producing a reconstruction error. This error drives the fine-tuning (i.e.\ TTT) of a small adaptor CNN in front of the segmentor, ultimately mapping the individual test images onto a normalised space that overcomes domain shifts problems for the segmentor. Our work achieves the same goal but uses GANs. Unlike TTANN, which separately pre-trains the mask DAE, our model is end-to-end because it can learn the shape prior while optimising the segmentor. Moreover, it reduces computational requirements as discriminators need less GPU memory and computation than autoencoders.

Concurrent with our work, \cite{he2021autoencoder} propose to use auto-encoders for TTT. They propose a bespoke model that can be fine-tuned at test time to adapt to local distributions using the decoder. However, our contribution is orthogonal to their work as \cite{he2021autoencoder} do not introduce any shape prior during adaptation.

Herein, we open new research directions towards learning re-usable discriminators that can improve segmentation performance at inference. We also show that we can further enhance test-time performance by combining the discriminator with reconstruction costs (via decoders), thus endowing a causal perspective to TTT.

\subsection{Tackling Distribution Shifts}
In recent years, improving model robustness under distribution shifts has attracted considerable attention in medical imaging, where images vary among scanners, patients, and acquisition protocols \citep{castro2020causality}.
In this context, domain adaptation and generalisation have become relevant research areas. Several methods attempt to learn domain invariant representations by anticipating the distribution difference between training and test \citep{joyce2017robust,li2018domain,dou2019domain,guan2021domain,zhou2021domain}. 
However, these approaches usually require prior knowledge about the test data, such as a small subset of (possibly labelled) images from the test distribution. 
Unfortunately, these data can be expensive or even impossible to acquire for every target domain, and distribution shifts might be not easily identifiable \citep{recht2018cifar}. 

An alternative approach is adapting the network parameters directly to the test samples \citep{sun2020test,karani2021test}. 
Similarly, our method does not need to simulate test distribution shifts, as it automatically adapts the segmentor to the individual test instances. Thus, our approach can be assumed to perform \textit{one-sample unsupervised domain adaptation} on the fly. Notice also that, compared to standard domain adaptation techniques, Test-time Training has the advantage that it does not become ill-defined when there is only one sample from the target domain and does not require the source data during test-time training.

\subsection{Shape Priors in Deep Learning for Medical Segmentation}
In the past, several methods have introduced shape priors into segmentation models \citep{nosrati2016incorporating, jurdia2020high} in the form of penalties \citep{kervadec2019constrained,clough2019topological,jurdi2021surprisingly}, atlases \citep{dalca2019unsupervised}, autoencoders \citep{oktay2017anatomically,dalca2018anatomical}, post-processing operations \citep{painchaud2019cardiac,larrazabal2020post}, 
and adversarial learning \citep{yi2019generative,valvano2021learning}.
Thanks to their flexibility, GANs are a popular way of introducing data-driven shape priors \citep{yi2019generative}, with the advantage of learning the prior while also optimising the segmentor.

\subsection{Re-using Adversarial Discriminators}
Pre-trained discriminators have been re-used to navigate the generator's latent space \cite{liu2020collaborative}, as anomaly detectors \citep{zenati2018adversarially,ngo2019fence}, or as features extractors for transfer learning \citep{radford2015unsupervised,donahue2016adversarial,mao2019discriminator}. 
To the best of our knowledge, no prior work uses them to detect test time segmentor mistakes or to fine-tune pre-trained segmentors at inference.

\section{Proposed Method}
Below, we first provide an overview of our method. Then, we describe the challenges of re-using adversarial discriminators at inference, suggesting possible solutions to address them. Lastly, we detail model architecture, training, and the re-use of discriminators at test time.

\textit{Notation.} We use capital Greek letters to denote functions $\mathfunc{f}$, italic lowercase for scalars $\mathscal{s}$, and bold lowercase for 2D images $\mathimage{x} \in \mathds{R}^{h \times w}$, being $h, w \in \mathbb{N}$ image height and width.

\subsection{Method Overview}\label{subsec:method_overview}
\begin{figure}[t]
    \centering
    \includegraphics[width=0.8\linewidth]{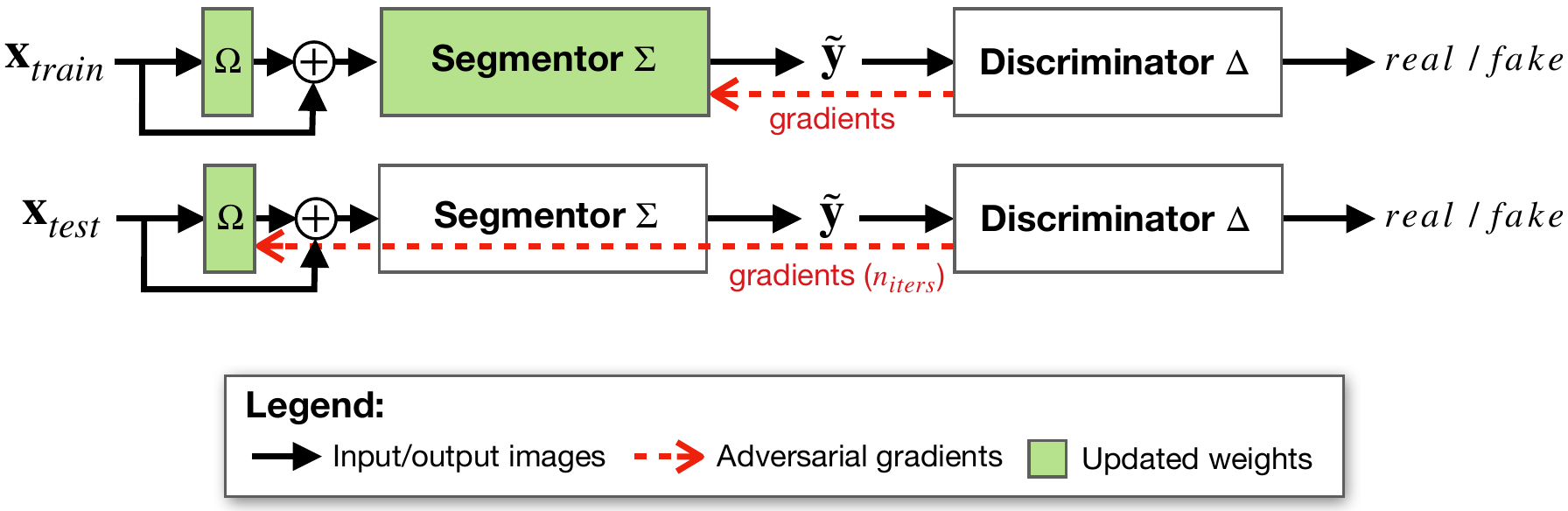}
    \caption{
    We re-use adversarial discriminators to correct segmentation mistakes at inference. As thoroughly discussed in Section~\ref{ch8:subsec:challenges_and_solutions}, crucial to the method success is training stable and re-usable discriminators. At inference, we tune a shallow adaptor $\mathfunc{w}$ on each test sample $\mathimage{x}$ independently, until predictions $\mathimage{\tilde{y}}$ satisfy the adversarial shape prior. We only need one sample for fine-tuning. 
    }
    \label{ch8:fig:method_overview}
\end{figure}

As we summarise in Fig.~\ref{ch8:fig:method_overview}, to segment an image we process it through a small adaptor $\mathfunc{w}$ and a subsequent segmentor $\mathfunc{s}$. When annotations are available, we train both the adaptor and segmentor to minimise a supervised cost. For unlabelled images, we instead optimise them based on an adversarial cost. Meanwhile, we train an adversarial discriminator to discern real and predicted segmentation masks.
At inference, we fine-tune the adaptor $\mathfunc{w}$ on each test sample, leveraging only the (unsupervised) adversarial loss to increase performance. 

Note that developing novel segmentors and adaptors is not our scope. Thus, we use previously developed architectures that have already shown success in segmentation tasks. 
On the contrary, a major contribution of this work is identifying the crucial challenges behind re-using adversarial discriminators at inference and suggesting possible solutions to overcome them.

\subsection{Re-usable Discriminators: Challenges and Solutions}\label{ch8:subsec:challenges_and_solutions}

\subsubsection{Challenge 1: Avoid Overfitting and Catastrophic Forgetting} Re-usable discriminators $\mathfunc{d}$ must not \textit{overfit} nor \textit{catastrophically forget}. Otherwise, their predictions at test time will not be reliable. Ensuring this condition holds is challenging because GANs can easily memorise data if trained for too long \citep{nagarajan2018theoretical}.\footnote{Notice that memorisation can also happen just in the discriminator. In fact, contrarily to the segmentors, there is no supervised cost to regularise the discriminator's training. We show how to detect memorisation from the losses in Appendix \hyperlink{appendix:a}{A}.
} Moreover, once the segmentor's training has converged \citep{shrivastava2017learning}, the discriminator may forget how unrealistic segmentation masks look like.
In these cases, although $\mathfunc{d}$ may work well at training, it would not generalise on test data, as we explain below.

At convergence, a properly trained segmentor $\mathfunc{s}$ predicts \textit{realistic} segmentation masks. Thus, we stop training of standard GANs while optimising $\mathfunc{d}$ to tell apart \textit{real} from more and more \textit{real-looking} masks. 
During the latest stages of training, this produces ambiguous training signals for the discriminator, which consistently receives real-looking inputs but is trained to label them as \textit{real} half the time, as \textit{fake} the other half. Since the discriminator loss will encourage $\mathfunc{d}$ to classify realistic masks as fake (even if it was the segmentor to generate them), the training gradients will become unreliable. At that point, the discriminator training becomes unstable, and $\mathfunc{d}$ collapses to one of these cases:
\textbf{i)} always predicting the equilibrium point (which in vanilla GANs is the number 0.5, equidistant from the labels \textit{real}: 1, \textit{fake}: 0) but  still able of detecting unrealistic images; 
\textbf{ii)} predicting the equilibrium point independently of the input image, forgetting what \textit{fake} samples look like \citep{shrivastava2017learning,kim2018memorization}; 
or \textbf{iii)} memorising the real masks (which, differently from the generated ones, appear unchanged since the beginning of training) and always classifying them as \textit{real}, while classifying \textit{any other input} as \textit{fake}. 
It is crucial to prevent the last two behaviours (\textbf{ii)} and \textbf{iii)}) to have a re-usable discriminator. Thus, we use:
\begin{itemize}
    \item \textit{Fake anchors}: We want to expose the discriminator to unrealistic masks (labelled as \textit{fake}) until the end of training. In particular, we train $\mathfunc{d}$ using real masks $\mathimage{y}$, predicted masks $\tilde{\mathimage{y}}$, and corrupted masks $\mathimage{y}_{corr}$. We obtain $\mathimage{y}_{corr}$ by randomly swapping squared patches within the image\footnote{We use patches having size equal to 10\% of the image size.} and adding binary noise to the real masks, as this proved to be a fast and effective strategy to learn robust shape priors in autoencoders \citep{karani2021test}.
    While, towards the end of the training, the discriminator may not distinguish $\mathimage{y}$ from the real-looking $\tilde{\mathimage{y}}$, the exposure to $\mathimage{y}_{corr}$ will prevent forgetting how unrealistic masks look like, ensuring informative training gradients.\footnote{Concurrent to our work, \cite{sinha2021negative} recently introduced a similar idea, named Negative Data Augmentation, which improved the training of GAN generators. However, differently from \cite{sinha2021negative}, we highlight that our scope is to build a stable discriminator, which can be re-used at inference.}
\end{itemize}

\subsubsection{Challenge 2: Ensuring Stability} An additional challenge is to train \textit{stable} discriminators, which do not change much during the last training epochs. In other words, we want to make oscillations of the discriminator loss as small as possible. This is necessary because we typically stop training using early stopping criteria on the segmentor loss. Therefore, we want to promote the optimisation of Lipschitz smooth discriminators, avoiding suddenly big gradient updates (which make the training loss oscillate).
\footnote{
Lipschitz smooth discriminators have better stability \citep{chu2020smoothness} and, thus, their training gradients are bounded. Hence: i) abrupt changes in their weight values are unlikely to happen, and ii) the discriminator behaviour does not change abruptly across epochs, which is crucial at convergence. In fact, by comparing different discriminators obtained between subsequent epochs (i.e. different statistical realization of the adversarial discriminator, having different weights), we find similar behaviours, reducing the influence of eschewing discriminator-based early stopping criteria.}

Hence, we limit the risk of having discriminators residing in sharp local minima, which are well known to generalise worse~\citep{keskar2016large}, using:
\begin{itemize}
    \item \textit{Smoothness constraints}: We use Spectral Normalisation \citep{miyato2018spectral}, \textit{tanh} activations, and Gradient Penalty \citep{wgangp} to encourage discriminator smoothness \citep{chu2020smoothness}.
    \item \textit{Discriminator data augmentation}: We use Instance Noise \citep{sonderby2016amortised,muller2019does} and random roto-translations, to map similar inputs to the same prediction label. We generate noise sampling from a Normal distribution having zero mean and 0.1 standard deviation. We rotate images between $0 \div \pi/2$ and translate them up to 10\% of image pixels on both vertical and horizontal axes. 
\end{itemize}

\subsection{Architectures and Training Objectives}\label{subsec:architectures_and_training}
Given an input image $\mathimage{x}$, we first pass it through the adaptor $\mathfunc{w}$ and obtain $\mathimage{x}'=\mathfunc{w}(\mathimage{x})$. Then, we use the segmentor $\mathfunc{s}$ to predict a segmentation mask $\tilde{\mathimage{y}}=\mathfunc{s}(\mathimage{x}')=\mathfunc{s}\circ\mathfunc{w}(\mathimage{x})$. 

We parametrised $\mathfunc{w}$ as the adaptor introduced by \cite{karani2021test}. Such an adaptor consists in 3 convolutional layers with 16 $3\times3$ kernels and activation $\mathfunc{f}(\mathtensor{t}) = e^{-\mathtensor{t}^2 / \mathscal{s}^2}$, where $\mathtensor{t}$ is an input tensor and $\mathscal{s}$ a trainable scaling parameter, randomly initialised and optimised at test-time. Instead, for the segmentor we use a UNet \citep{ronneberger2015u} with batch normalisation \citep{ioffe2015batch}.

For the pairs of annotated data ($\mathimage{x}$, $\mathimage{y}$) in the training set, we minimise the weighted cross-entropy loss between the real mask $\mathimage{y}$ and the predicted one $\tilde{\mathimage{y}}$: 
\begin{align}\label{ch8:eq:sup_loss}
        \mathcal{L}(\mathfunc{w}, \mathfunc{s})  
                = - \sum\nolimits_{\mathscal{i}=1}^{\mathscal{c}} 
                \mathscal{w}_i \cdot \mathimage{y}_i \log(\tilde{\mathimage{y}}_i),
\end{align}
where, given the number of classes $\mathscal{c}$ and the class index $i$, we address the class imbalance problem with the scaling factor $\mathscal{w}_i$. We compute $\mathscal{w}_i = 1 - \mathscal{n}_i/\mathscal{n}_{tot}$ as the ratio between the number of pixels $n_i$ having label $\mathscal{i}$ and the total number of pixels $\mathscal{n}_{tot}$.

The discriminator $\mathfunc{d}$ is a convolutional encoder, processing an input mask through 5 convolutional layers with $4 \times 4$ filters. The number of filters follows the series: 32, 64, 128, 256, 512. After the first two layers, we use a stride of 2 to downsample the extracted features maps.
As discussed in Section~\ref{ch8:subsec:challenges_and_solutions}, we increase discriminator smoothness through spectral normalisation layers and \textit{tanh} activations. 
Lastly, we use a fully-connected layer to integrate high-level representation and produce a scalar linear output, used to compute the adversarial loss adapted from \cite{mao2018effectiveness}:
\begin{equation}\label{ch8:eq:adv_loss}
\begin{split}
         & \min_{\mathfunc{d}} \bigg\{
         \mathcal{V}_{LS}(\mathfunc{d}) = 
            \frac{1}{2} E_{\mathimage{y} \sim p(\mathimage{y})}[(\mathfunc{d}(\mathimage{y}) - 1)^2] 
            + 
            \frac{1}{2} E_{\mathimage{x} \sim p(\mathimage{x})}[(\mathfunc{d}(\mathfunc{s}\circ\mathfunc{w}(\mathimage{x})) + 1)^2] \bigg\}
        \\ &
        \min_{\mathfunc{w}, \mathfunc{s}} \bigg\{
        \mathcal{V}_{LS}(\mathfunc{w}, \mathfunc{s}) = 
            \frac{1}{2} E_{\mathimage{x} \sim p(\mathimage{x})}[(\mathfunc{d}(\mathfunc{s}\circ\mathfunc{w}(\mathimage{x})))^2] \bigg\},
\end{split}
\end{equation}
where $+1$ and $-1$ are the labels for \textit{real} and \textit{fake} masks, respectively.
For training, we alternately minimise Eq.~\ref{ch8:eq:sup_loss} on a batch of labelled images and Eq.~\ref{ch8:eq:adv_loss} on a batch of unpaired images and unpaired masks. To avoid the adversarial loss from prevailing over the supervised cost, we rescale $\mathcal{V}_{LS}(\mathfunc{w}, \mathfunc{s})$ by multiplying it by a dynamic weighting value $\mathscal{a}=0.1 \cdot \frac{\norm{\mathcal{L}(\mathfunc{w}, \mathfunc{s})}}{\norm{\mathcal{V}_{LS}(\mathfunc{w}, \mathfunc{s})}}$, as in \cite{valvano2021learning}. 
Hence, we ensure to expose the segmentor to a supervised cost that is always one order of magnitude larger than the adversarial cost, which can judge predictions only qualitatively. 
We minimise losses with Adam \citep{kingma2014adam}, learning rate: $10^{-4}$, and batch size: 12. We end training when the supervised loss stops decreasing on a validation set.

\subsection{Adversarial Test-Time Training}\label{ch8:subsec:adapting_omega}
\subsubsection{Adapting $\mathfunc{w}$}
At test time, we only fine-tune the first few convolutional layers of the whole model: i.e. the three layers of the adaptor $\mathfunc{w}$ described in Section \ref{subsec:architectures_and_training}. Our choice is motivated by \cite{asano2019critical} observations that the early layers are the most suited for one-shot learning, and is similar to that of \cite{karani2021test}. Leaving the deeper layers of $\mathfunc{s}$ unchanged, we let the model adapt only to changes at lower abstraction levels, limiting its flexibility and preventing trivial solutions.
Thus, given a test sample $\mathimage{x}$, we tune a shallow convolutional residual block (the adaptor $\mathfunc{w}$) in front of the segmentor by minimising $\mathcal{V}_{LS}(\mathfunc{w}|\mathfunc{s}, \mathimage{x})$ for $\mathscal{n}_{iter}$ iterations. The number of iterations $\mathscal{n}_{iter}$ has an upper bound and is determined on each specific test sample independently. 
After tuning $\mathfunc{w}$, the input to the segmentor becomes an augmented version of $\mathimage{x}$, which can be more easily classified. 

\subsubsection{Setting the Number of Test-time Iterations}
\label{ch8:subsec:exp_computational_aspects}
At inference, our method needs $\mathscal{n}_{iter}$ forward and backward passes to correct a segmentation.\footnote{Despite being slower than standard inference, where each image only requires one forward pass, we observe just a small temporal overhead in the model: $\sim$10$\div$20s/patient on a TITAN Xp GPU.}
We compute an optimal $\mathscal{n}_{iter}$ for each test sample, stopping TTT when the adversarial loss %
on the predicted mask has not decreased for the last $200$ steps, or when TTT exceeds a maximum iteration number $\mathscal{n}_{iter}^{max}=1000$.
After stopping TTT, we consider the prediction associated with the minimum adversarial loss as the best one.



\section{Experimental Setup}\label{ch8:sec:experimental_setup}

\subsection{Data}\label{ch8:subsec:data}
We consider four medical datasets acquired using a variety of MRI scanners and acquisition protocols: cardiac data of ACDC \citep{bernard2018deep}, M\&Ms \citep{mnms}, and LVSC \citep{suinesiaputra2014collaborative}; and the abdominal organ data of CHAOS \citep{kavur2021chaos}. 
ACDC and M\&Ms contain annotations for right ventricle, left ventricle and left myocardium, while LVSC only has myocardium masks. CHAOS contains labels for the two kidneys, the liver, and the spleen.
We use specific datasets based on two different learning scenarios, which we describe below:
\begin{itemize}
    \item \textbf{Identifiable Distribution Shift}: We use ACDC and M\&Ms data to model test-time distribution shifts that we can identify as changes in the MRI acquisition scanner. 
    For ACDC, we build the training and validation set using only data acquired from 1.5T scanners; then, we test the model on 3T MRI scans. In the following, we refer to this dataset setup as ACDC\textsubscript{1.5\textrightarrow 3T}. 
    For M\&Ms, we consider training and validation sets containing data from 3 out of the four available MRI vendors and a test set constructed using data from the held-out vendor. As a result, we can be sure that there is a distribution shift between training and test data in both ACDC\textsubscript{1.5\textrightarrow 3T} and M\&Ms. In both cases, we maintain a 2:1 ratio between the number of samples in the training and validation sets.
    
    \item \textbf{Non-identifiable Distribution Shift}: We consider randomly sampled data from ACDC, LVSC, and CHAOS, where we cannot say in advance if there is a change in distribution between train and test data. We consider a semi-supervised learning scenario, where only a portion of training data is annotated. This setup is of particular interest when collecting large-scale annotated datasets is not possible, and small labelled training sets do not accurately represent the test distribution. To prevent information leakage, we divide datasets by patients and use groups of 40\%-20\%-40\% of patients for training, validation, and test set, respectively. In ACDC and LVSC training sets, we only consider annotations for one fourth of the training subjects (10 patients); in CHAOS, only one half (4 patients). We treat the remaining data as unpaired and use them for adversarial learning (Eq.~\ref{ch8:eq:adv_loss}). Despite being drawn from the same distribution (i.e.\ the entire dataset), the small amount of training data may not fully represent the data distribution. Hence, although we cannot identify distribution shifts a priori, they may still exist and lead to performance drop \citep{recht2018cifar}. 
\end{itemize}

After defining train, validation and test sets, we pre-process data as follows:
\begin{itemize}
    \item \textbf{ACDC\textsubscript{1.5\textrightarrow 3T}} and \textbf{ACDC:} we resample images to the average resolution of 1.51$mm^2$, and crop or pad them to $224\times224$ pixel size. Lastly, we normalise data by subtracting the patient-specific median and dividing by its interquartile range (IQR).
    \item \textbf{M\&Ms:} after resampling the images to the average resolution of 1.25$mm^2$, we crop/pad them to $224\times224$ pixels. We normalise images by subtracting the patient-specific median and dividing by the IQR.
    \item \textbf{LVSC:} we resample images to the average resolution of 1.45$mm^2$, and then crop or pad them to $224\times224$ pixel size. Finally, we normalise images by subtracting the patient-specific median and dividing by the IQR.
    \item \textbf{CHAOS:} we test our method on the T1 in-phase images, after resampling them to 1.89$mm^2$, normalising and cropping them to $192\times192$ pixel size. 
\end{itemize}

\subsection{Evaluation Protocol}
For all the experiments, we report results of 3-fold cross-validation. We measure performance in terms of segmentation quality, using Dice score, IoU score, and Hausdorff distance to compare the predicted segmentation masks with the ground truth labels available in the test sets. We assess statistical significance with the bootstrapped t-test. We use significance at $p \leq 0.05$ or $p \leq 0.01$ denoted by one (*) or two (**) asterisks, respectively.

\section{Experiments and Discussion}\label{ch8:sec:experiments_and_discussion}
We present and discuss the performance of our method in various experimental settings. 
First, Section~\ref{ch8:subsec:exp_attt} presents the advantage of the proposed approach during inference: either under identifiable or non-identifiable distribution shifts. 
In Section~\ref{ch8:subsec:exp_postproc}, we highlight the differences between adversarial TTT and post-processing operations, reporting also complementary performance gains. 
After that, Section~\ref{ch8:subsec:exp_online_learning} shows model potential for Online Continual Learning.
Lastly, Section~\ref{ch8:subsec:causal_ttt} discusses a limitation of the approach and defines a possible solution to overcome it building on a causal perspective: we show that including losses on image self-reconstruction during Test-time Training makes adaptation more reliable and faster. 

\begin{figure}
    \centering
    \includegraphics{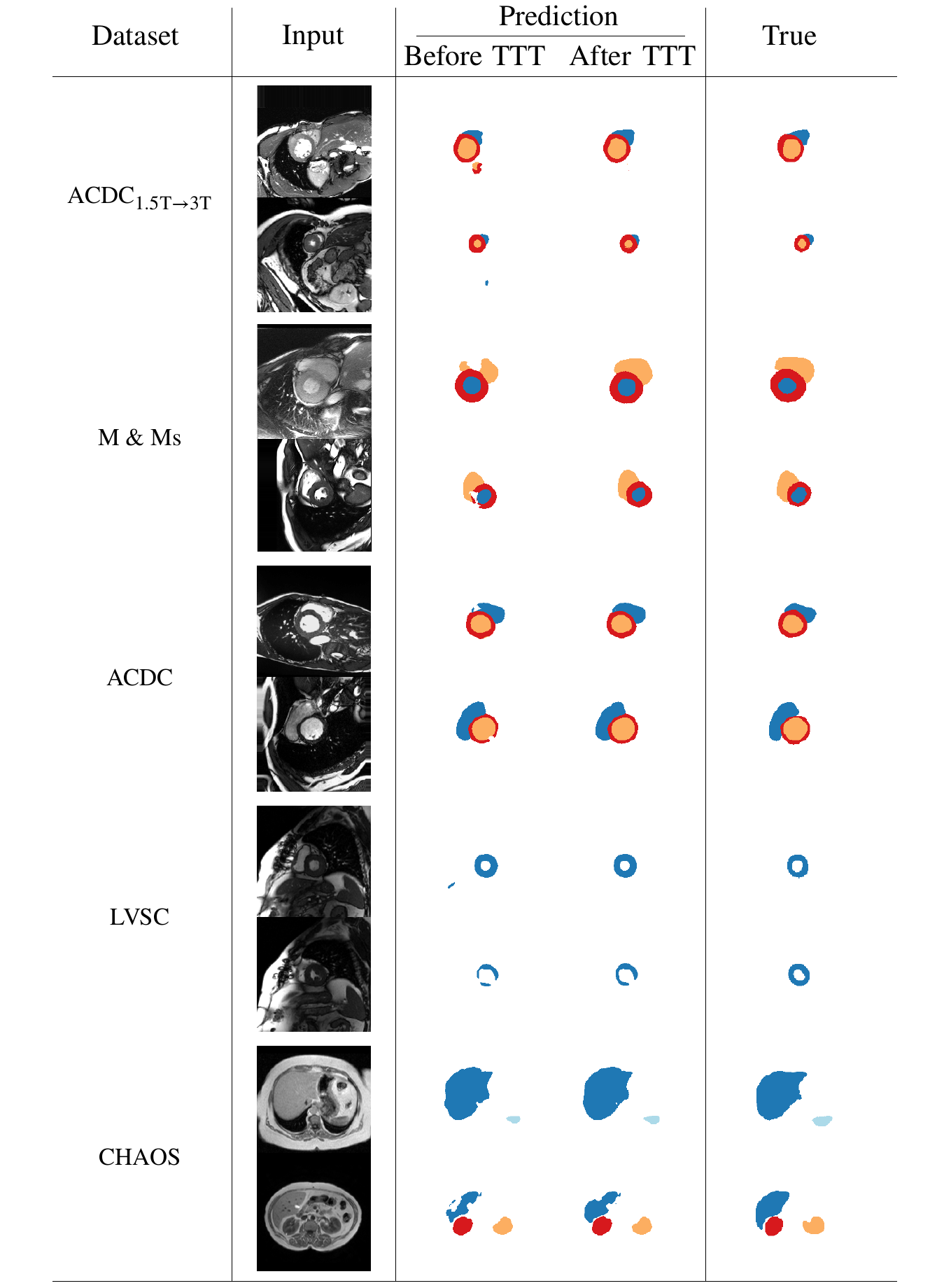}
    \caption{
    We show examples of prediction mistakes and their corrections after the adversarial Test-time Training. We group pairs of examples by dataset. As can be observed, the segmentor corrects the initially er\-ro\-ne\-ous segmentation masks to make them realistic, according to the learned adversarial shape prior.
    }
    \label{ch8:fig:adaptation_and_segmentation}
\end{figure}


\begin{table}
    \centering
        \begin{tabular}{l |c |c|c|c}
        	\multirow{2}{*}{Dataset} & 
        	\multirow{2}{*}{Adv. TTT} & 
        	\multirow{2}{*}{Dice ($\uparrow$)} & 
        	\multirow{2}{*}{IoU ($\uparrow$)} & 
        	Hausdorff \multirow{2}{*}{($\downarrow$)} \\
        	&&&& Distance ~~~~~\\
        	\midrule
        	
            \multirow{2}{*}{ACDC\textsubscript{1.5\textrightarrow 3T}} 
            & 
            before &  
            77.0\textsubscript{09}~~~ & 
            68.9\textsubscript{09}~~~ & 
            5.2\textsubscript{02}~~ \\
            & 
            \hcell{after} & 
            \hcell{\textbf{78.4\textsubscript{08}}**} &  \hcell{\textbf{70.4\textsubscript{08}}**} & \hcell{\textbf{5.0\textsubscript{02}}**}\\
        	
        	\midrule
        	
            \multirow{2}{*}{M\&Ms}
            & 
            before & 
            82.0\textsubscript{08}~~~ & 
            75.6\textsubscript{08}~~~ & 
            4.4\textsubscript{03}~~ \\
            & 
            \hcell{after} & 
            \hcell{\textbf{82.1\textsubscript{08}}*~} &  \hcell{\textbf{75.7\textsubscript{08}}*~} & \hcell{\textbf{4.3\textsubscript{03}}~~~}\\
        	
        	\midrule
        	
            \multirow{2}{*}{ACDC}
            &
            before &  
            74.2\textsubscript{10}~~~ & 
            66.1\textsubscript{10}~~~ & 
            7.1\textsubscript{06}~~ \\
            & 
            \hcell{after} & 
            \hcell{\textbf{75.0\textsubscript{09}}**} &  \hcell{\textbf{67.1\textsubscript{10}}**} & \hcell{\textbf{6.9\textsubscript{05}}~~~}\\
        	
        	\midrule
        	
            \multirow{2}{*}{CHAOS}
            & 
            before & 
            74.0\textsubscript{12}~~~ & 
            70.3\textsubscript{12}~~~ & 
            9.1\textsubscript{04}~~ \\
            & 
            \hcell{after} & 
            \hcell{\textbf{74.3\textsubscript{12}}**} &  \hcell{\textbf{70.5\textsubscript{12}}**} & \hcell{9.1\textsubscript{04}~~~}\\
            
        	\midrule
        	
            \multirow{2}{*}{LVSC}
            & 
            before &  
            62.6\textsubscript{15}~~~ & 
            53.1\textsubscript{14}~~~ & 
            5.8\textsubscript{04}~~ \\
            & 
            \hcell{after} & 
            \hcell{\textbf{65.9\textsubscript{12}}**} &  \hcell{\textbf{56.2\textsubscript{12}}**} & \hcell{\textbf{5.7\textsubscript{03}}~~~}\\
        	
        	\bottomrule
        \end{tabular}
    \caption{Dice (\textuparrow), IoU (\textuparrow) and Hausdorff distance (\textdownarrow) obtained before and \hltext{after} tuning the segmentor on the individual test instances. Arrows show metric improvement direction; numbers are the average performance, with standard deviation as subscript; best results are in \textbf{bold}. 
    Observe how adversarial Test-time Training always improves performance (bootstrapped t-test, $^{*}p<0.05$, $^{**}p<0.01$). }
    \label{ch8:tab:before_after_ttt}
\end{table}

\subsection{Adversarial Test-time Training Under Distribution Shifts}\label{ch8:subsec:exp_attt}

We start with a qualitative example of test-time adaptation in Fig.~\ref{ch8:fig:adaptation_and_segmentation}, showing that it helps fix prediction mistakes. As can be seen on all datasets, our method corrects unrealistic masks by removing scattered false positives and segmentation holes.

In Table~\ref{ch8:tab:before_after_ttt}, we report segmentation performance before and after Test-time Training.
We find performance improvements across metrics and datasets both in terms of metric average and spread. The only case where differences are not statistically significant is on CHAOS data, where the test set has a small number of samples (8 patients), and distributions are broad. Nevertheless, we observe empirical improvements in terms of Dice and IoU scores on CHAOS, too.
From these results, we argue that adversarial TTT could lead to substantial benefits for medical applications, where systems must be robust and avoid trivial mistakes.

\begin{figure}
    \centering
    \includegraphics[width=\linewidth]{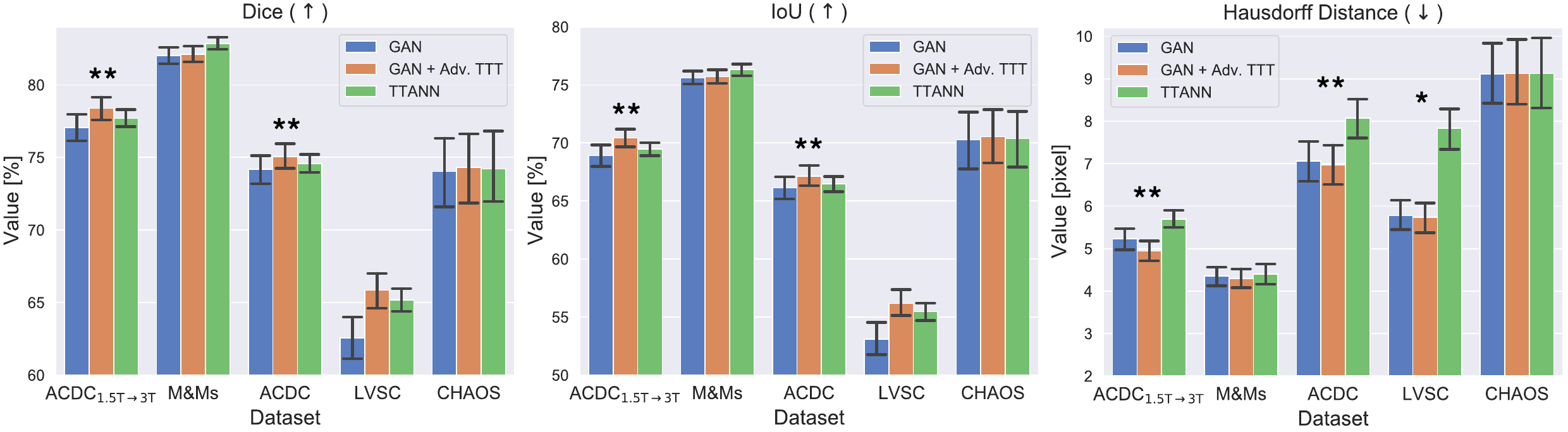}
    \caption{
    Adversarial TTT has competitive performance with TTA\-NN \citep{karani2021test}, and it has the advantage of re-using an already available GAN component. Bar plots report average performance and standard errors. Stars on top of the bar plots show if differences between adversarial TTT and TTANN are statistically significant (bootstrapped t-test, $^{*}p<0.05$, $^{**}p<0.01$).
    }
    \label{ch8:fig:bar_plot_ttan}
\end{figure}

In Fig.~\ref{ch8:fig:bar_plot_ttan}, we compare our method with one using a shape prior separately learned by a DAE (i.e.\ TTANN, \cite{karani2021test}). 
We compare the performance increases obtained through adversarial TTT vs using TTANN, and discuss the pros and cons of driving the adaptation using a mask discriminator vs a DAE.
Our experiments show advantages in using our method. Although performance gains appear small and TTANN performs better on M\&Ms data, using adversarial TTT leads to statistically significant improvements in most of the cases. 
Probably, the performance increase derives from the optimisation procedure, as we train the discriminator to detect the segmentor mistakes. On the contrary, DAEs are optimised independently of the segmentor by only artificially simulating prediction mistakes. 
Thus, DAEs may have never seen specific mask corruptions during training, as also observed by \cite{larrazabal2020post}. 


\textit{Ablation Study:} In Table~\ref{ch8:tab:ablation_acdc} we show results ablating the adaptor, the smoothness constraints and the proposed \textit{fake anchors} regularisation on the model. As shown, the techniques improve training and make the adversarial shape prior stronger. Consequently: i) the adversarial loss trains a better segmentor, and ii) the re-usable discriminator can increase test-time performance. For comparison, training a simple UNet on the same data leads to an average Dice score of 70.1 (standard deviation of 13).
\begin{table}
    \centering
        \begin{tabular}{l |c|c| c|c|c}
            \multirow{2}{*}{} & \multirow{2}{*}{Adaptor $\mathfunc{w}$} & Smoothness & Fake & Adversarial & \multirow{2}{*}{Performance}\\
        	& & Constraints & Anchors & TTT & \\
        	\midrule
        	\hcell{Ours} & \hcell{\checkmark} & \hcell{\checkmark} & \hcell{\checkmark}& \hcell{\checkmark}
        	& \hcell{75.0\textsubscript{09}} \\
        		
        	\#1 & \checkmark & \checkmark & \checkmark & ~ & 74.2\textsubscript{10} \\
        	\#2 & \checkmark & \checkmark & ~ & ~ & 72.8\textsubscript{12} \\
        	\#3 & \checkmark & ~ & ~ & ~ & 72.7\textsubscript{12} \\
        	\#4 & ~ & ~ & ~ & ~ & 70.0\textsubscript{12} \\
         		                
        	\bottomrule
        \end{tabular}
    \caption{Ablation Study. We compare the performance of our method (\hltext{Ours}) after removing: adversarial Test-time Training (ablation \#1), the proposed regularisation technique (\textit{fake anchors}, \#2), the smoothness constraints discussed in Section~\ref{ch8:subsec:challenges_and_solutions} (ablation \#3), and the adaptor (standard GAN, \#4). Performance is in terms of average Dice score on ACDC data, with standard deviation as subscript.}
    \label{ch8:tab:ablation_acdc}
\end{table}
We also report an ablation study comparing the contributions of the each type of fake anchors regulariser in Table \ref{tab:ablation_fake_anchors}. In this experiment, we trained the model using no fake anchors regulariser, only patch swap, only binary noise, or both. After training, we evaluated the segmentor on the test set before TTT. As shown in the table, both randomly swapping segmentation patches and adding binary noise contribute to train better segmentors, and TTT improves upon it further.

\begin{figure}[t]
    \centering
    \includegraphics[width=0.9\linewidth]{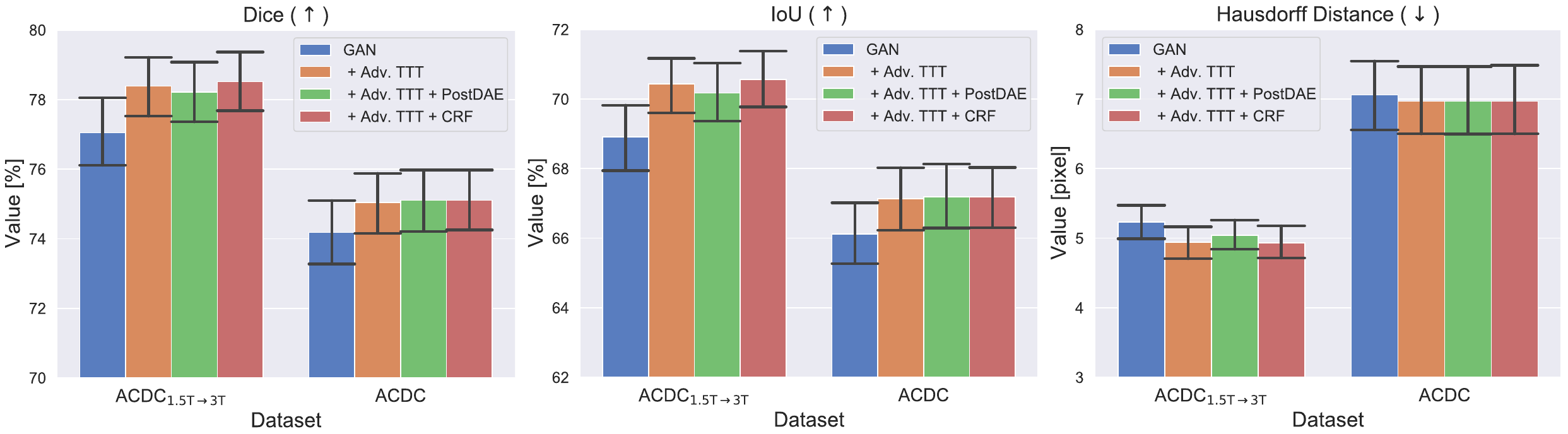}
    \caption{Compatibility with post-processing techniques (PostDAE and CRF). Bar plots report average performance and standard errors.
    }
    \label{ch8:fig:bar_plots_post_processing}
\end{figure}

\begin{table}
    \centering
        \begin{tabular}{l| c|c|c|c|c}
            & None & Patch Swap & Binary Noise & Both & Both + Adversarial TTT \\
        	\midrule
        	
        	Performance 
        	& 72.8\textsubscript{12}
        	& 73.2\textsubscript{11}
        	& 72.9\textsubscript{10}
        	& 74.2\textsubscript{10}
        	& 75.0\textsubscript{09}
        	\\
         		                
        	\bottomrule
        \end{tabular}
    \caption{Ablation study on the fake anchors regularisers. Performance are in terms of average Dice score, with standard deviation as subscript, on the ACDC test set. Both patch swapping and adding binary noise contribute to train better segmentors.}
    \label{tab:ablation_fake_anchors}
\end{table}

\subsection{Combining Adversarial TTT with Post-processing}\label{ch8:subsec:exp_postproc}

Adversarial TTT should not be confused with post-processing operations because it does not modify the prediction independent of the input. Indeed, our approach lets the model adapt to the input image.
Moreover, contrary to standard post-processing, our method has the advantage that it can also learn from a continuous stream of data, as we will show in the next section. However, this does not mean that post-processing cannot follow after test-time training, which we now explore.

As examples, we consider two popular post-processing techniques. First, we examine post-processing with Conditional Random Fields (CRF), as in the DeepLab framework \citep{chen2017deeplab}. Note that CRF adapts the \textit{predicted mask} to the image, assigning nearby pixels with similar colours to the same semantic class. Our method, instead, adapts the \textit{model} to the image. Second, we consider correcting the segmentation mistakes with a Denoising Autoencoder, as in PostDAE \citep{larrazabal2020post}. This method maps corrupted masks on a previously learned manifold of realistic masks without considering the associated images. 

Fig.~\ref{ch8:fig:bar_plots_post_processing} shows that our method can be combined with both techniques and, sometimes, the combination can improve performance. We find that PostDAE does not always help: probably because adversarial TTT already adapts the model using a data-driven shape prior (via the discriminator) and the additional DAE may be uninformative or even harmful. On the contrary, CRF increases performance because it introduces a different type of prior in the model \citep{zheng2015conditional}, from which the segmentor can benefit (similar to what happens in model ensembling).


\subsection{Online Continual Learning of Adversarial TTT}\label{ch8:subsec:exp_online_learning}
We now experiment with the possibility of using our method for Online Continual Learning \citep{delange2021continual,mai2021online}, i.e.\ learning from a continuous stream of non-stationary data (in our case, data affected by distribution shifts). 
Learning from new data, the model performance should gradually increase. Moreover, as the model gets better on the test distribution, the need for TTT should decrease, making Test-time Training faster. 

\begin{figure}[t]
    \centering
    \includegraphics[width=0.8\linewidth]{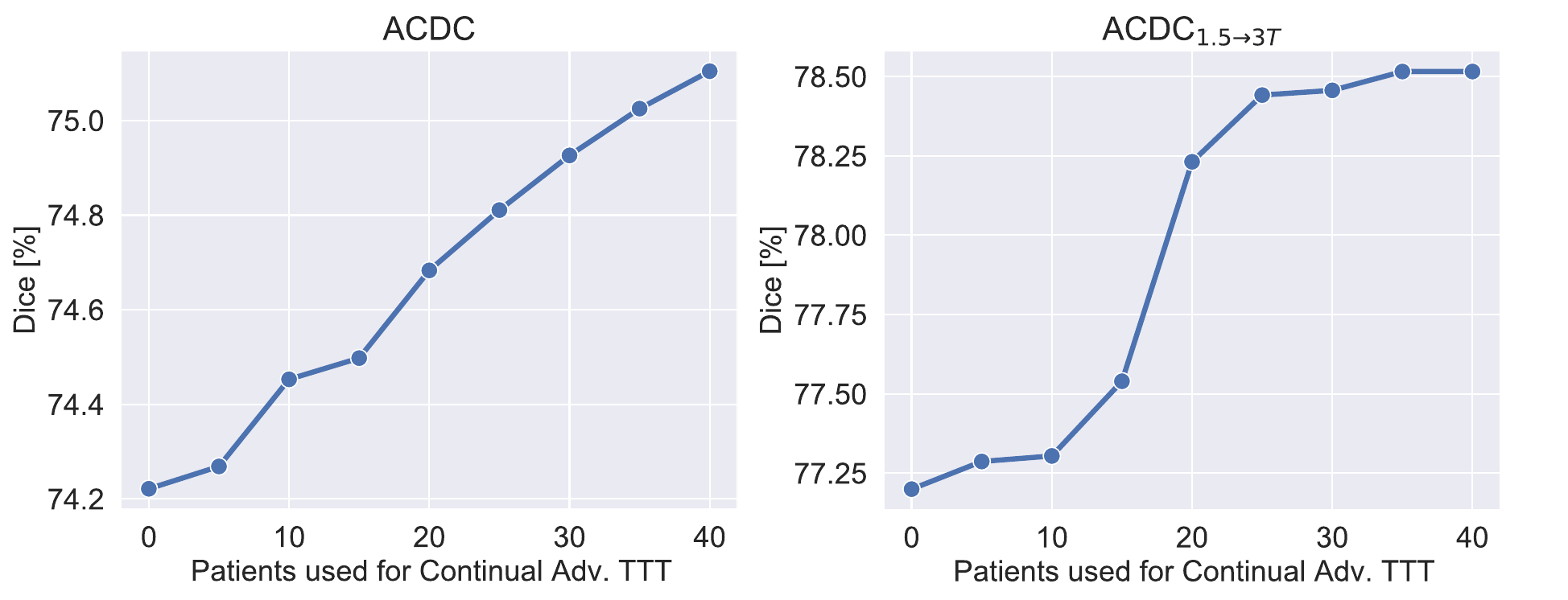}
    \caption{
    Effect of increasing the number of patients used to adapt the model in Online Continual Learning. For this experiment, we continually adapt the model on $k$ test patients, and then perform standard inference on the remaining test patients. We span $k$ between 0 (i.e. no adaptation) and 40 (i.e. continual learning on the whole test set). As new data become available, the model adapts to new distribution shifts and improves overall test performance.
    }
    \label{fig:continual_learning}
\end{figure}

\begin{table}
    \centering
        \begin{tabular}{l |c|c |c|c|c}
        	\multirow{2}{*}{Dataset} & 
        	\multirow{2}{*}{Adv. TTT} & 
        	\multirow{2}{*}{Continual} & 
        	\multirow{2}{*}{Dice ($\uparrow$)} & 
        	\multirow{2}{*}{IoU ($\uparrow$)} & 
        	Hausdorff \multirow{2}{*}{($\downarrow$)}\\
        	&&&&& Distance ~~~~~\\
        	\midrule
        	
            \multirow{3}{*}{ACDC} 
            & \xmark & \xmark &  74.2\textsubscript{10} & 66.1\textsubscript{10} & 7.1\textsubscript{05} \\
            & \checkmark & \xmark &  75.0\textsubscript{09} & 67.1\textsubscript{10} & 6.9\textsubscript{05}\\
            & \hcell{\checkmark} & \hcell{\checkmark} &  \hcell{\textbf{75.1\textsubscript{09}}} &  \hcell{\textbf{67.2\textsubscript{09}}} & \hcell{\textbf{6.9\textsubscript{05}}}\\
        	
        	\midrule
        	
            \multirow{3}{*}{ACDC\textsubscript{1.5\textrightarrow 3T}}
            & \xmark & \xmark &  77.0\textsubscript{09} & 68.9\textsubscript{09} &  5.2\textsubscript{02} \\
            & \checkmark & \xmark & 78.4\textsubscript{08} &  70.4\textsubscript{08} & 5.0\textsubscript{02}\\
            & \hcell{\checkmark} & \hcell{\checkmark} &  \hcell{\textbf{78.6\textsubscript{08}}} &  \hcell{\textbf{70.6\textsubscript{08}}} &  \hcell{\textbf{4.9\textsubscript{02}}}\\
        	
        	\bottomrule
        \end{tabular}
    \caption{Online Continual Learning. Our model can continuously learn from a stream of test data, gradually improving segmentation performance. Numbers are average performance, with standard deviation as subscript. Best results in \textbf{bold}.}
    \label{ch8:tab:continual_learning}
\end{table}

We conduct experiments for both ACDC and ACDC\textsubscript{1.5\textrightarrow 3T}, and report results in Figure~\ref{fig:continual_learning} and Table \ref{ch8:tab:continual_learning}. 
In this continual learning scenario, we do not restart TTT from zero when testing new data, but we continue the learning process from one patient in the test set to another. For each patient we perform the adaptation only once.

Overall, we find that the segmentor benefits from the continuous stream of test data, with performance increasing from one patient to another (Fig.~\ref{fig:continual_learning}), achieving even higher scores than using adversarial TTT on each test subject separately (Table~\ref{ch8:tab:continual_learning}). 

More interestingly, we find that the average number of TTT steps needed for tuning the adaptor in Online Continual Learning decreases from 322 to 315 on ACDC data and from 120 to 114 on ACDC\textsubscript{1.5\textrightarrow 3T}. This reduced number of steps suggests that gradually introducing new knowledge into the model lessens the need for adaptation, and the segmentor might be able to do without TTT after a while.


%
\subsection{Towards Causal Test-time Training}\label{ch8:subsec:causal_ttt}
\begin{figure}[t]
    \centering
    \includegraphics[width=0.65\linewidth]{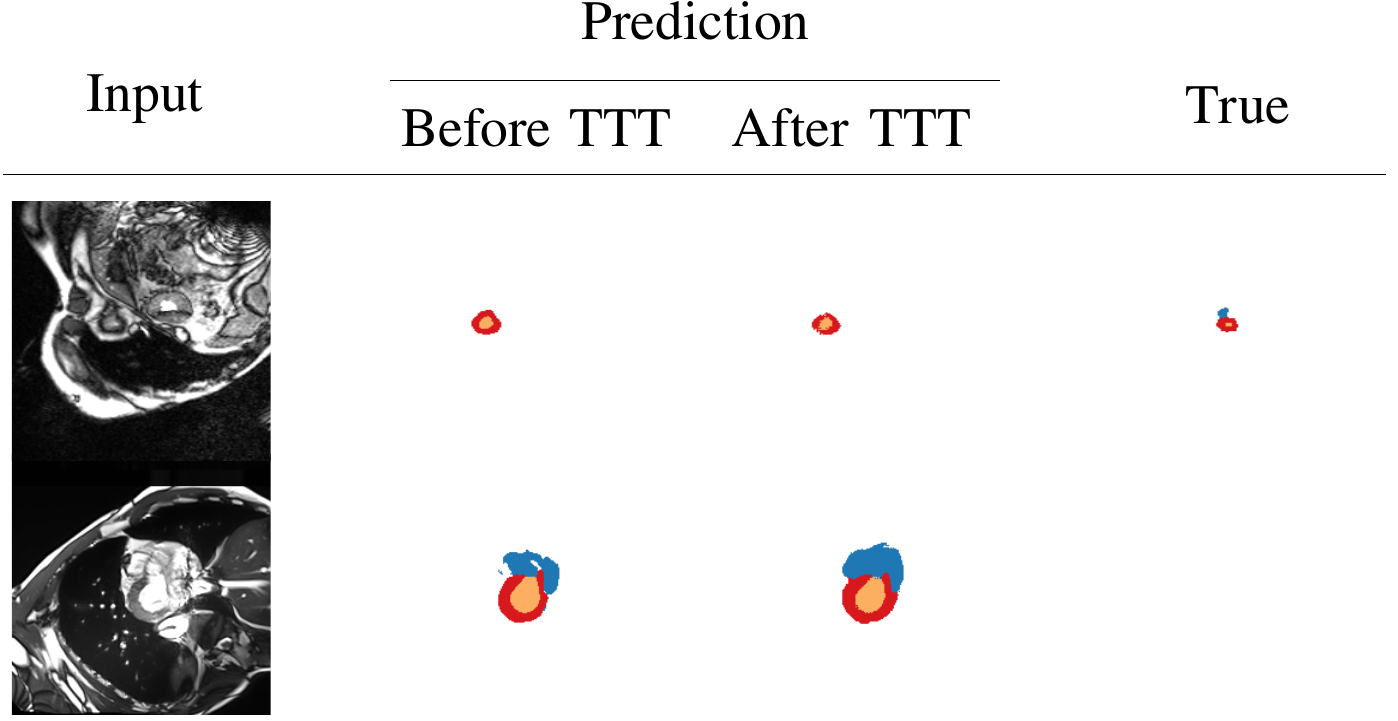}
    \caption{
    Since the information contained in the predicted mask is limited, the discriminator does not penalise realistic but wrong predictions (\textbf{top row}). Sometimes, it may even encourage to make bigger mistakes (\textbf{bottom row}).
    }
    \label{ch8:fig:shape_prior_mistakes}
\end{figure}

We experimentally observed that, in few and rare cases, adversarial TTT can make segmentation worse (see such an example in Fig.~\ref{ch8:fig:shape_prior_mistakes}). This happens because the discriminator learns to approximate the shape prior characterised by the probability distribution $p(\mathimage{y})$ rather than the joint distribution $p(\mathimage{x},\mathimage{y})$. 
Thus, the discriminator will not penalise a realistic mask even when it is the wrong segmentation for the given image (see Fig.~\ref{ch8:fig:shape_prior_mistakes}). Hence, it is natural to wonder whether considering both the image and the segmentation mask to drive the adaptation process may provide additional context and help Test-time Training. 
We emphasise that this problem also exists in TTANN \citep{karani2021test} and in all the methods learning the marginal $p(\mathimage{y})$ instead of the joint distribution of images and masks. 

As we discuss in Appendix \hyperlink{appendix:b}{B}, we can give a causal interpretation to this problem, highlighting that image-related information would make Test-time Training causal and more effective. Below, we investigate if we can include such information and how it affects TTT.

\begin{figure}[t]
    \centering
    \includegraphics[width=0.9\linewidth]{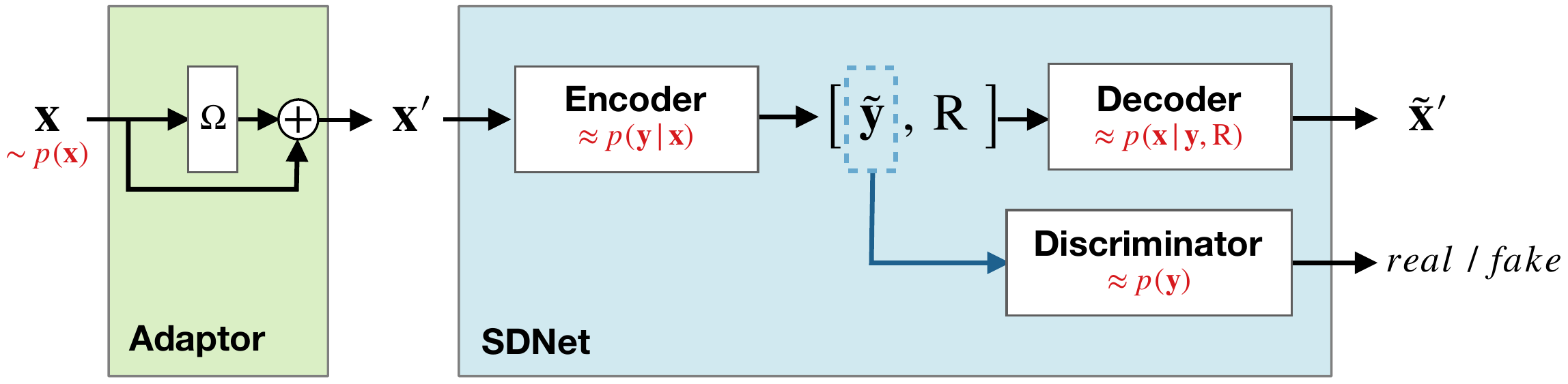}
    \caption{
    The proposed approach in a causal setting. We add the adaptor $\Omega$ in front of SDNet \citep{chartsias2019disentangled} to transform an image $\mathimage{x}\sim p(\mathimage{x})$ into its adapted version $\mathimage{x}'$. \textbf{During training}, the SDNet encoder extracts the segmentation mask $\tilde{\mathimage{y}}$ and a residual representation $\mathtensor{R}$. A decoder uses both of them to reconstruct the adapted image, predicting $\tilde{\mathimage{x}}' \approx \mathimage{x}'$. Meanwhile, a mask discriminator learns to tell apart real segmentation masks from the predicted ones. \textbf{At inference}, we perform Test-time Training and adapt $\Omega$ to minimise the sum of the reconstruction cost (computed comparing $\mathimage{x}'$ and $\tilde{\mathimage{x}}'$) and the adversarial loss (computed on the predicted $\tilde{\mathimage{y}}$ according to Eq.~\ref{ch8:eq:adv_loss}).
    }
    \label{ch8:fig:causal_ttt_schematic}
\end{figure}

For our experiments, we consider a method designed for semi-supervised segmentation learning: SDNet \citep{chartsias2019disentangled}, which we describe schematically in Fig.~\ref{ch8:fig:causal_ttt_schematic}. 
SDNet disentangles images using an encoder to find latent representations regarding the image content (anatomy) and style (modality), and a decoder to reconstruct the input images. Task specific networks learn to predict segmentations from the content latent, using either supervised losses or adversarial costs.  
As such, SDNet learns the joint distribution between images and masks (for additional details: see Appendix \hyperlink{appendix:b}{B}), making SDNet a perfect test bench for evaluating both the use of discriminators and image reconstruction for TTT.

We explore if reconstructing the test sample at inference adds benefits during adaptation in terms of \textit{performance} and \textit{adaptation speed}. 
We include an adaptor $\mathfunc{w}$ in front of SDNet (as shown in Fig.~\ref{ch8:fig:causal_ttt_schematic}). Then, we train the full model according to the SDNet training objectives, which include an adversarial loss $\mathcal{L}_A$ and a reconstruction loss $\mathcal{L}_R$.

The adversarial loss $\mathcal{L}_A$ is the same as defined in Eq.~\ref{ch8:eq:adv_loss}, and we also follow the precautions discussed in Section~\ref{ch8:subsec:challenges_and_solutions}. We leave the reconstruction term $\mathcal{L}_R$ as in the original SDNet framework, to minimise the mean absolute error between an image and its reconstruction. However, we train the model to reconstruct the adapted image $\mathimage{x}'=\mathfunc{w}(\mathimage{x})$ rather than the input $\mathimage{x}$. We motivate this specific change by observing that under a distribution shift between training and inference data, the SDNet decoder may not be able to reconstruct the test image correctly. Instead, after tuning $\mathfunc{w}$ to the test image, the SDNet can effectively reconstruct the adapted image $\mathimage{x}'$.\footnote{An alternative to reconstructing $\mathimage{x}'$ would be to introduce an ``inverted'' adaptor $\mathfunc{w}^{-1}$ at the decoder output. However, this would require extra computational cost and reconstructing $\mathimage{x}'$ is simpler.}
We leave the rest of the SDNet model unchanged.

During inference, we fix the SDNet weights, and do Test-time Training to tune the adaptor on each sample. 
We set the number of TTT steps $\mathscal{n}_{iter}$ as described in Section~\ref{ch8:subsec:adapting_omega}, using the adaptation loss described in three different settings:
\begin{itemize}
    \item ``SDNet + Rec. TTT'', where we do TTT using only the reconstruction loss $\mathcal{L}_R$;
    \item ``SDNet + Adv. TTT'', where we drive adaptation using only the adversarial loss $\mathcal{L}_A$;
    \item ``SDNet + Adv. \& Rec. TTT'', where we use the sum of the adversarial and the reconstruction cost $\mathcal{L}_{tot} = \mathcal{L}_A + \mathcal{L}_R$, leading to a consistent causal-driven adaptation.
\end{itemize}

\begin{figure}[t]
    \centering
    \includegraphics[width=\linewidth]{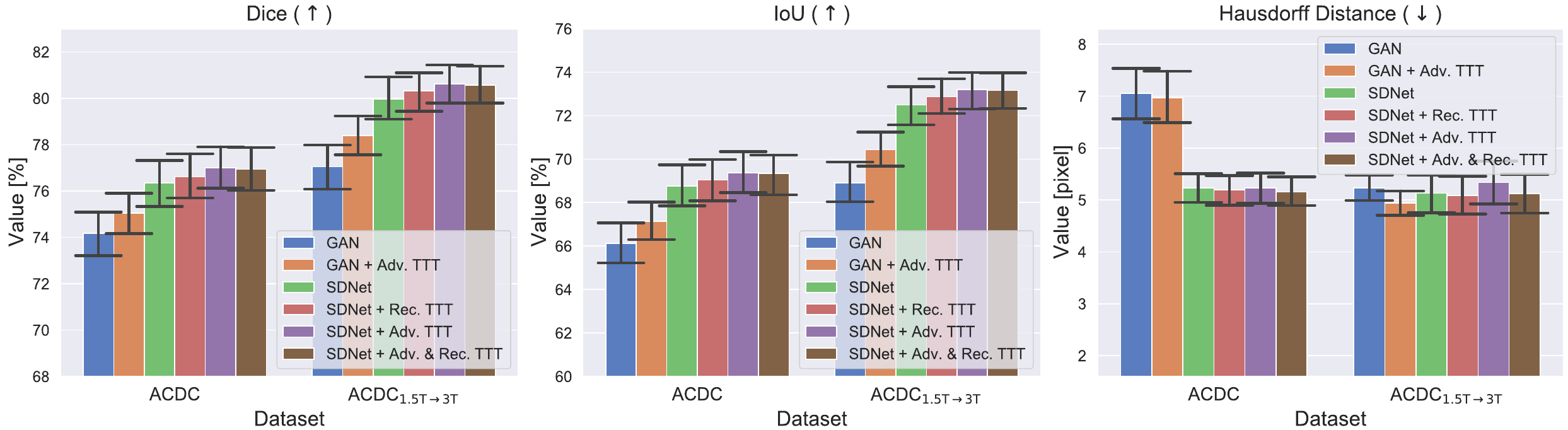}
    \caption{
    We compare the performance of: a GAN before and after adversarial Test-time Training; the SDNet model (discussed in Section~\ref{ch8:subsec:causal_ttt}); the SDNet after Test-time Training performed minimising only a reconstruction cost (``+ Rec. TTT''), only an adversarial cost (``+ Adv. TTT'') and their sum (``+ Adv. \& Rec. TTT''). Bar plots report average performance and standard errors. 
    }
    \label{ch8:fig:bar_plots_sdnet_ttt}
\end{figure}

For the experiments, we considered both the case of clearly identifiable distribution shifts (ACDC\textsubscript{1.5\textrightarrow 3T} data) and non-identifiable shifts (ACDC data). We report per-dataset results in terms of segmentation quality in Fig.~\ref{ch8:fig:bar_plots_sdnet_ttt}. 

From this figure, we observe that all three types of TTT improve SDNet performance, confirming that the framework is general and widely applicable. In fact, both the adversarial discriminator and the decoder used to reconstruct the image provide useful priors to drive the adaptation process. 
There is only one experimental exception: the Hausdorff distance of ``SDNet + Adv. TTT'' on ACDC\textsubscript{1.5\textrightarrow 3T} data. 
In this case, while Dice and IoU scores increase, the Hausdorff distance  worsens. 
We believe this happens because ``SDNet + Adv. TTT'' makes more errors in the most apical and basal slices of the heart\footnote{By definition, the Hausdorff distance between two binary masks has the maximum possible value (i.e.\ the image size) when one of the two masks is empty. In this case, even one missed or one extra pixel in the apical and basal slices leads to high values of the metric, decreasing performance.}: a behaviour that we also observe for ``GAN'' and ``GAN + Adv. TTT'', where Hausdorff distances are high.

Analysing the contribution of the reconstruction loss in detail, we observe that it mainly helps when optimising the model on the training data (i.e.\ before Test-time Training). In fact, if we compare the performance of SDNet with that of a GAN \textit{before} TTT (``SDNet'' vs ``GAN'', in Fig.~\ref{ch8:fig:bar_plots_sdnet_ttt}), we find a big improvement in all the metrics. On the contrary, when we analyse the effect of $\mathcal{L}_R$ during Test-time Training, we find that it only slightly affects the metrics (compare ``SDNet + Adv. TTT'' vs ``SDNet + Adv. \& Rec. TTT''). 

Instead, including $\mathcal{L}_R$ during TTT has a bigger impact on the test-time training speed. In fact, we find that the number of TTT iterations needed for convergence halves. Specifically, using only the adversarial cost during TTT, the average optimal $\mathscal{n}_{iter}$ is 111 on ACDC, and 206 on ACDC\textsubscript{1.5\textrightarrow 3T} data. By adding also the reconstruction term, the average number of TTT steps becomes 66 on ACDC and 119 on ACDC\textsubscript{1.5\textrightarrow 3T}. Our results are also in line with recent findings arguing that correct causal structures adapt faster \citep{bengio2019meta}.

These experiments highlight that causal TTT, using the causal structure herein, leads to marginal improvements in segmentation quality, but it makes adaptation to the test samples considerably faster. 


\section{Conclusion}
In this work, we suggest re-using adversarial discriminators at inference. First, we identify simple design assumptions that must be satisfied to make discriminators useful once training is complete. Then, we show how to use discriminators to detect and correct segmentation mistakes on the test data. 
The proposed approach is simple. compatible with common post-processing techniques, and it increases test-time performance on the most challenging images. 
Our approach also benefits from continual learning, making test-time inference gradually more accurate and faster. 
Lastly, we showed that reconstruction losses could complement mask discriminators and improve the inference speed of our model.

We believe that learning without supervision on new test data is a promising research avenue. However, while this works only touches upon the potential use of discriminators and their combination with other components at inference time, there are still several challenges to solve. 
First, Test-time Training methods rely on gradient descent to fine-tune the model on the test data. While effective, this solution slows down model prediction time. 
Hence, future work should focus on strategies to make TTT faster, increasing parallelisation and possibly adapting the model using just a single training step. 
For example, \cite{bartler2021mt3} have recently explored the combination of Meta-Learning techniques with TTT, to make adaptation faster. 
Furthermore, when moving towards continual learning, alleviating the risk of forgetting previous experience is crucial. To this end, it would be interesting to combine our method with approaches aiming to solve these challenges, such as Elastic Weight Consolidation \citep{Kirkpatrick3521}.

More generally, re-using adversarial discriminators to fix generator mistakes may open several research opportunities, even outside image segmentation. 
Thanks to their flexibility and the ability to learn data-driven losses, GANs have been widely adopted in medical imaging, from domain adaptation to image synthesis tasks \citep{yi2019generative}. In this context, we believe that improved architectures and regularisation techniques \citep{kurach2019large, chu2020smoothness} will make adversarial networks even more popular. 
Thus, training stable and re-usable discriminators opens opportunities for re-using flexible data-driven losses at test time and make inference better. 


\acks{This work was partially supported by the Alan Turing Institute (EPSRC grant EP/N510129/1). S.A. Tsaftaris acknowledges the support of Canon Medical and the Royal Academy of Engineering and the Research Chairs and Senior Research Fellowships scheme (grant RCSRF1819\textbackslash8\textbackslash25). We thank NVIDIA for donating the GPU used for this research.}

%
\ethics{The work follows appropriate ethical standards in conducting research and writing the manuscript, following all applicable laws and regulations regarding treatment of animals or human subjects.}

\coi{We declare we don't have conflicts of interest.}

\bibliography{references}

\newpage

\appendix 

\section*{Appendix \hypertarget{appendix:a}{A.} Discriminator: Convergence and Memorisation}




During adversarial training, we optimise the segmentor to produce realistic masks, which the discriminator cannot distinguish from the real ones. At convergence, the discriminator can either reach an equilibrium stage or collapse.

Below, we show empirical examples of the training and validation losses for the GAN discriminator $\mathfunc{d}$. For these experiments, we use a Least-square GAN \citep{mao2018effectiveness}, whose discriminator loss to minimise is:
\begin{equation}\label{eq:discriminator_loss}
     \mathcal{V}_{LS}(\mathfunc{d}) = 
        \frac{1}{2} \underbrace{
            E_{\mathimage{y} \sim \mathdistrib{Y}}[(\mathfunc{d}(\mathimage{y}) - 1)^2]
        }_{\text{loss on \textit{real} samples}}
        + 
        \frac{1}{2} \underbrace{
            E_{\mathimage{x} \sim \mathdistrib{X}}[(\mathfunc{d}(\mathfunc{s}(\mathimage{x})) + 1)^2] 
        }_{\text{loss on \textit{fake} samples}}
\end{equation}
where $+1$ and $-1$ are the labels for \textit{real} and \textit{fake} (generated) images, respectively.

\begin{figure}[h]
    \centering
    \includegraphics[width=\linewidth]{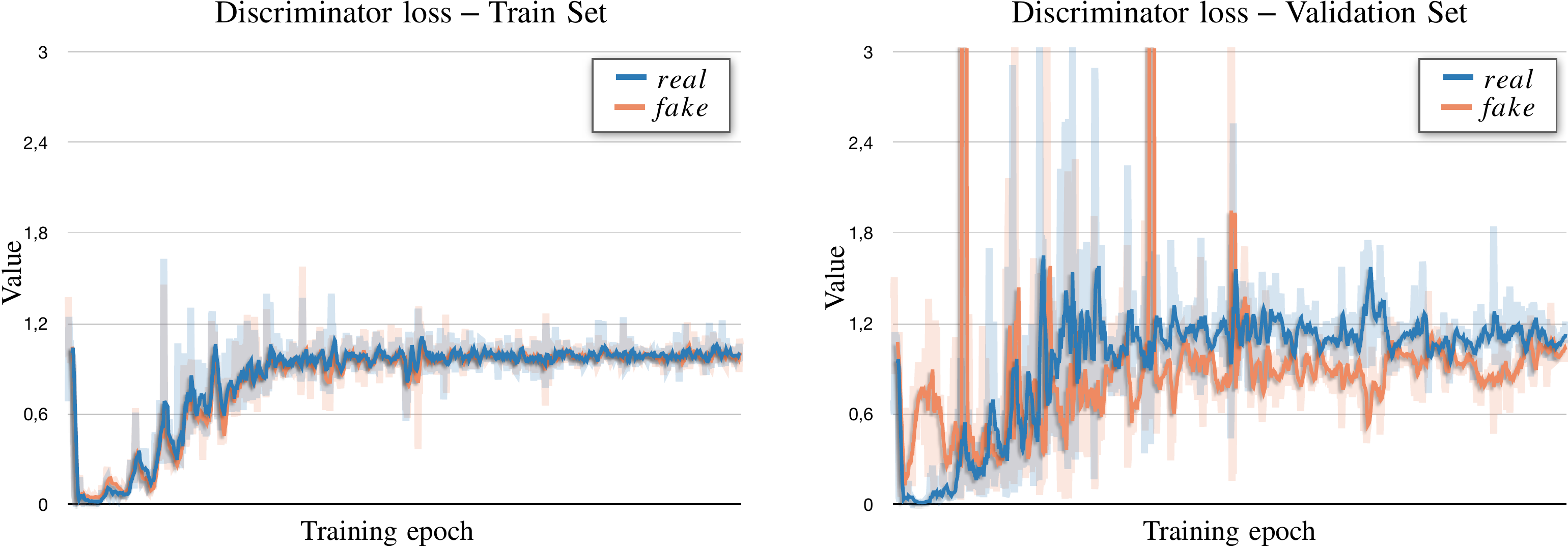}
    \caption{
    At convergence, the discriminator reaches an equilibrium where it always predicts the value 0, equidistant from the \textit{real} and the \textit{fake} labels. Thus, losses for \textit{real} and \textit{fake} masks converge to the equilibrium value 1.0 in both train and validation.
    }
    \label{fig:gan_collapse_equilibrium}
\end{figure}

\begin{figure}
    \centering
    \includegraphics[width=\linewidth]{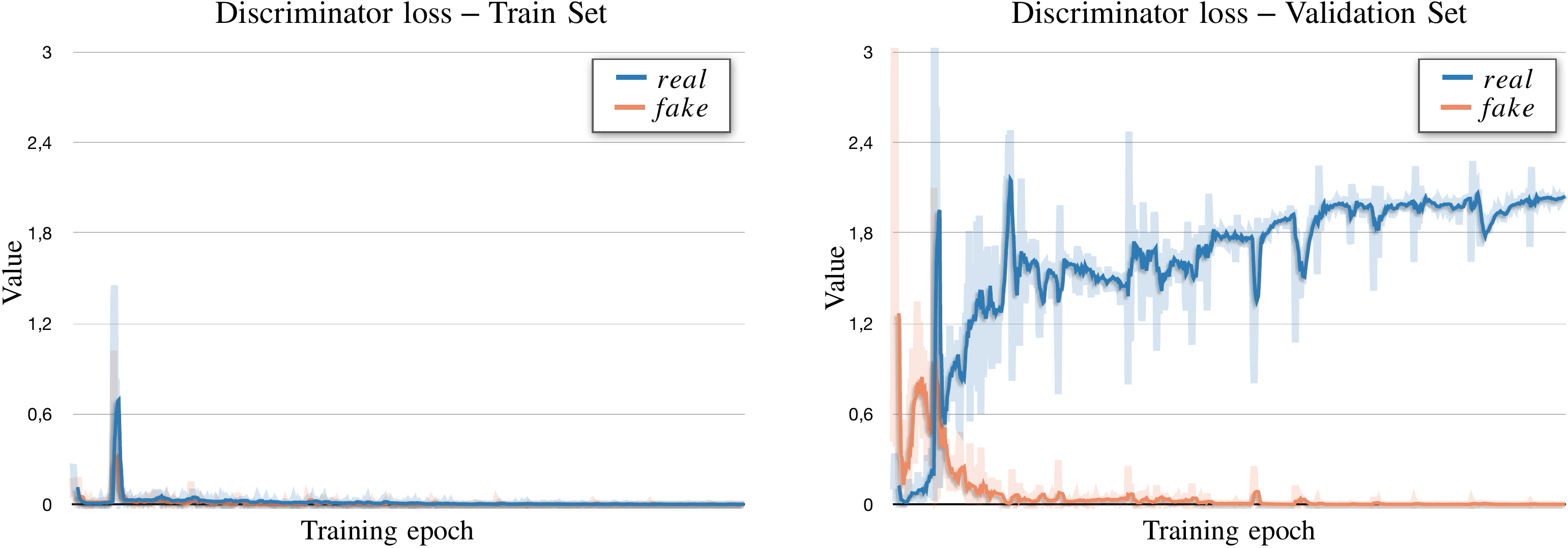}
    \caption{
    At convergence, the discriminator shows signals of memorisation. The discriminator memorises the \textit{real} training images, and it predicts the label \textit{fake} (i.e. the value -1) for any other case. During validation, the \textit{fake} images are still classified correctly, while the \textit{real} ones are classified as \textit{fake} and the associated loss converges to the value of 2.0. 
    }
    \label{fig:gan_collapse_memorisation}
\end{figure}

We report examples of convergence modes in Fig.~\ref{fig:gan_collapse_equilibrium} and Fig.~\ref{fig:gan_collapse_memorisation}. On the left side: losses on the training set; on the right: losses on the validation set. Observe that $-$ despite the single loss components have different values $-$ the total loss $\mathcal{V}_{LS}(\mathfunc{d})$ on the validation set is the same in both cases.

\section*{Appendix \hypertarget{appendix:b}{B.} Causal Test-time Training}




Causal machine learning is recently gaining considerable attention in medical imaging \citep{castro2020causality} because it could identify the best suited approaches to solve a specific task \citep{scholkopf2013semi,castro2020causality}, or make learning faster \citep{bengio2019meta}.

In our model, we optimise the segmentation modules of a GAN ($\mathfunc{w}$ and $\mathfunc{s}$, described in Section \ref{subsec:method_overview}) to approximate the conditional distribution $p(\mathimage{y}|\mathimage{x})$, and the discriminator to learn the marginal $p(\mathimage{y})$. Training this type of GAN has practical advantages because the discriminator regularises the segmentor and allows to use unlabelled data for training. 
However, tuning the adaptor based only on the adversarial loss may be considered as driving the adaptation using an anti-causal model, while the process is instead causal. In other terms, it is an image that causes the predicted mask because experts draw masks on top of the images, and not vice versa \citep{castro2020causality}. Instead, GANs whose discriminator only receives segmentation masks as input would penalise the segmentor without considering the image causing that mask.

Thus, from a causal perspective, our approach is non-optimal. To capture the causal structure better and update the model parameters, we should also consider the inverse conditional probability $p(\mathimage{x}|\mathimage{y})$, which would improve the approximation $p(\mathimage{y}|\mathimage{x})$ (i.e. the segmentation modules $\mathfunc{w}$ and $\mathfunc{s}$) according to Bayes theorem:
\begin{equation}\label{ch8:eq:bayes}
    p(\mathimage{y}|\mathimage{x}) = p(\mathimage{x}|\mathimage{y}) \frac{p(\mathimage{y})}{p(\mathimage{x})}.
\end{equation}

Hence, to obtain a more coherent description, we should learn an inverse function mapping the masks to their respective images: $p(\mathimage{x}|\mathimage{y})$. Unfortunately, segmentation masks do not contain all the information needed to go from $\mathimage{y}$ to $\mathimage{x}$, as one mask can be associated to many different images, also known as the \textit{one-to-many problem}. 
Since this inversion is not possible, rather than learning the two components $p(\mathimage{x}|\mathimage{y})$ and $p(\mathimage{y})$ separately (Eq.~\ref{ch8:eq:bayes}), one may attempt to directly learn the joint distribution $p(\mathimage{x}, \mathimage{y})=p(\mathimage{x}|\mathimage{y})p(\mathimage{y})$.
To have this type of model, we can optimise the discriminator $\mathfunc{d}$ providing input pairs $(\mathimage{x}, \mathimage{y})$ rather than just unpaired masks. 
As a result, $\mathfunc{d}$ would implicitly learn to approximate the joint probability distribution of image-mask pairs, rather than the distribution of masks, and we would obtain a coherent causal description.
However, this approach also has several problems. 
In the first place, the discriminator would be subject to pixel-intensity distribution shifts of $\mathimage{x}$: thus, we would move our problem from adapting $\mathfunc{w}$ to the test images, to the problem of adapting $\mathfunc{d}$. 
Moreover, since the discriminator would need paired data for training, we would not be able to use the framework in semi-supervised settings where we have unpaired images and unpaired segmentation masks (such as in the scenarios of non-identifiable distribution shift, described in Section~\ref{ch8:subsec:data}).\footnote{For completeness, we also conducted an experiment in fully-supervised learning, where all the training images are associated with a segmentation mask and where the discriminator can learn the joint distribution. In this case, we observed that the discriminator was more prone to overfitting the training data, and its generalisation under distribution shifts got worse.}

Another alternative to learning $p(\mathimage{x}|\mathimage{y})$ is to substitute it with a proxy distribution. For example, we could learn $p(\mathimage{x}|\mathimage{y}, \mathtensor{R})$, where $\mathtensor{R}$ is a residual representation containing complementary information that is not present in $\mathimage{y}$ and is necessary to go from a mask $\mathimage{y}$ to the respective image $\mathimage{x}$, breaking the one-to-many/many-to-one problem described above. In this case, we would establish the relationship:
\begin{equation}
    p(\mathimage{y}|\mathimage{x}) \leftrightarrow p(\mathimage{x}|\mathimage{y}, \mathtensor{R}) \frac{p(\mathimage{y})}{p(\mathimage{x})}.
\end{equation}

As discussed in the paper, an example of such a model is SDNet \citep{chartsias2019disentangled}, which uses the extracted mask and its residuals to reconstruct the image\footnote{To be more precise, this holds assuming that the ``anatomy encoder" of SDNet performs a segmentation task and extracts $\mathimage{y}$ within the anatomical representation of a patient. For the purpose of our experiments, we assume it is a reasonable approximation.}, while also having an adversarial discriminator learning $p(\mathimage{y})$.

\end{document}